\documentclass[letterpaper, 10pt, conference]{ieeeconf}
\IEEEoverridecommandlockouts  

\usepackage{cite}
\usepackage{amsmath,amssymb,amsfonts}
\usepackage{graphicx}
\usepackage{booktabs}
\usepackage{multirow}
\usepackage{todonotes}
\usepackage{float}

\usepackage{tikz}
\newcommand{\xmark}{%
\tikz[scale=0.23] {
    \draw[line width=0.7,line cap=round] (0,0) to [bend left=6] (1,1);
    \draw[line width=0.7,line cap=round] (0.2,0.95) to [bend right=3] (0.8,0.05);
}}

\usepackage{algorithm}
\usepackage{algpseudocode}

\usepackage[pagebackref,breaklinks,colorlinks]{hyperref}

\usepackage[capitalize]{cleveref}
\crefname{section}{Sec.}{Secs.}
\Crefname{section}{Section}{Sections}
\Crefname{table}{Table}{Tables}
\crefname{table}{Tab.}{Tabs.}

\begin{document}

\title{Vision-language Models for Driver Monitoring Systems: A Driver Activity Description Dataset}

\author{David J. Lerch$^{1,3 \star}$ \thanks{$^{\star}$ These authors contributed equally to this work.} \thanks{$^{1}$ Fraunhofer IOSB, Karlsruhe, Germany {\tt\small \{firstname.lastname\}@iosb.fraunhofer.de}}, Sarath Mulugurthi$^{2 \star}$ \thanks{$^{2}$ Technische Hochschule Ingolstadt, Ingolstadt, Germany}, Manuel Martin$^{1}$, Frederik Diederichs$^{1}$, Rainer Stiefelhagen$^{3}$ \thanks{$^{3}$ Karlsruhe Institute of Technology (KIT), Karlsruhe, Germany {\tt\small \{firstname.lastname\}@kit.edu}}}

\maketitle

\begin{abstract}
Understanding subtle driver actions is essential for building reliable driver monitoring systems. Existing vision-language models (VLMs) are trained on general datasets and struggle to recognize fine distinctions in driver behaviors. This paper addresses this limitation by creating a detailed natural language version of the Drive\&Act dataset. 
We evaluate three VLMs on our new benchmark using LLM-based scoring methods. Their performance on the new benchmark shows that they cannot reliably generate accurate fine-grained driver activity descriptions. Based on the labeled Drive\&Act dataset we create a new Drive\&Act description dataset containing fine-grained descriptions to train VLMs on driver activity understanding.
Cross dataset evaluation on the Driver Monitoring Dataset~(DMD) shows that the VLM fine-tuned on our new Drive\&Act description dataset generalizes well to actions in the DMD dataset. The VLM fine-tuned on our Drive\&Act description dataset achieves an ACCR score of 76 outperforming the zero-shot VLM baseline with an ACCR score of 66. These findings demonstrate that adapting VLMs with richly described driver actions can significantly improve their ability to interpret driver behavior while also highlighting the need for more diverse datasets to support broader generalization in future applications. Our Drive\&Act description dataset and code will be publicly available on GitHub.
\end{abstract}
\section{Introduction}
\label{sec:intro}
Driver monitoring systems~(DMS) have emerged as a critical technology for modern intelligent vehicles because human error, particularly inattention, distraction, and drowsiness, remains a dominant cause of road crashes and fatalities~\cite{EuropeanCommission2020VisionZero}. According to the EuroNCAP Vision 2030~\cite{EuroNCAP2022} the requirements for DMS continue to increase. In addition to safety systems driver activity recognition can also contribute to comfort e.g. to prevent motion sickness~\cite{Diederichs2024Activities}.
Within this context, driver activity recognition focuses on identifying what the driver is doing, such as using a phone, reaching for objects, adjusting controls, or taking hands off the wheel, in order to enable timely warnings, adapt automation behavior, and support shared control between human and vehicle.
Video-based and multi-modal activity recognition methods are therefore increasingly used to detect distracted, impaired, or unsafe driving behaviors~\cite{canas2021visigrapp, Hasan2024DriveCLIP}. 

\begin{figure}[t]
  \centering
  \includegraphics[width=0.9\linewidth]{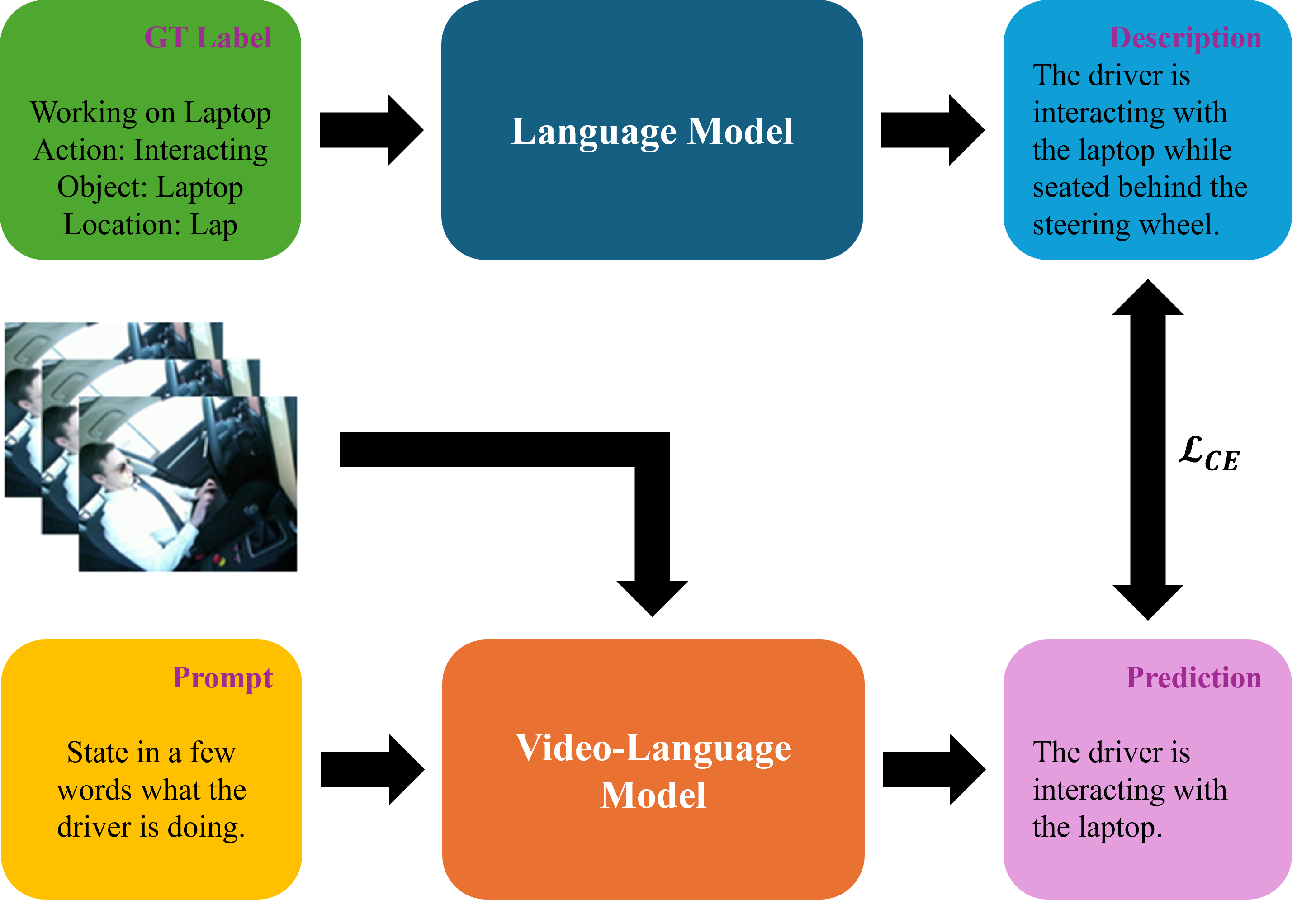}
   \caption{Overview over the pipeline of automatic description generation for driver actions. In the first step we use a labeled dataset and a LLM to generate ground truth descriptions. Thereafter we train a video-language model to predict the description based on a prompt and video input.}
   \label{fig:overview}
\end{figure}

While conventional deep models can classify predefined activities~\cite{xing2019dar}, they struggle with recognizing subtle distinctions between semantically similar fine-grained actions e.g. opening and closing a bottle~\cite{martin2019drive_and_act_2019_iccv}. This limitation is increasingly problematic as vehicles become more automated and in-cabin activities diversify, creating long-tail behaviors and domain shifts that are difficult to cover with fixed taxonomies and closed-set classifiers. Furthermore, interpretability is imperative in safety-critical automotive systems. For instance, controllers and human operators must comprehend not only the class label but also the rationale underlying the system's belief that a driver is distracted or engaged in a specific secondary task.
Vision-language models (VLMs) are promising for driver activity recognition because they can align visual signals with expressive natural language, enabling open-vocabulary descriptions and queries about driver behavior rather than restricting inference to a small set of predefined classes. By grounding video segments in textual descriptions, VLMs can represent the internal structure of activities as sequences of fine-grained events, which is well aligned with hierarchical driver behavior annotations such as atomic action triplets in datasets like Drive\&Act~\cite{martin2019drive_and_act_2019_iccv}. This expressiveness allows a single model to recognize both seen and novel driver behaviors via language prompts, improving generalization, which is particularly important as new in-cabin interfaces, devices, and interaction patterns appear over a vehicle’s lifetime.
Parameter-efficient fine-tuning strategies (e.g. LoRA) further make it feasible to specialize large pre-trained VLMs to the driver monitoring domain.
Consequently, integrating VLMs with fine-grained, temporally aligned driver activity annotations offers a compelling path toward more accurate, robust, and semantically meaningful driver monitoring systems that better support the safety objectives of future intelligent vehicles.

We show that current VLMs fail in describing driver activities. In order to deploy VLMs in DMS, driver activity datasets are required that enable effective training of such models~\cite{Canas2025ExplorationVLMsDMS}.
This paper proposes a pipeline (see Figure~\ref{fig:overview}) to train VLMs on driver activity recognition for DMS. Our main contributions are:
\begin{enumerate}
    \item \textbf{Benchmarking VLMs on Driver Activity Recognition:} We benchmark three SotA VLMs on the downstream task of driver activity recognition. We show that SotA VLMs fail in the downstream task of driver activity recognition underlining the necessity for finetuning and consequently training datasets.
    \item \textbf{New Driver Action Description Dataset:} We create a new dataset based on the existing Drive\&Act dataset containing fine-grained activity descriptions. Fine-grained descriptions are necessary for VLM training datasets.
    \item \textbf{Cross-dataset Generalization Capabilities:} We demonstrate that a VLM trained on our Drive\&Act-Description dataset can generalize to the DMD dataset~\cite{ortega2020dmd}. This finding underscores the necessity of fine-grained description datasets for in-cabin monitoring.
\end{enumerate}

\section{Related Work}
\label{sec:related_work}
\subsection{Vision-language Models}
VLMs were developed to bridge the gap between visual perception and linguistic understanding. 
The earliest efforts toward combining vision and language centered primarily on image captioning systems. These models relied on convolutional neural networks (CNNs) to extract visual features, which were then fed into recurrent neural networks such as LSTM-based decoders to generate textual descriptions. Examples include Show and Tell~\cite{vinyalis2015show}, which coupled a CNN encoder with an LSTM decoder, and Show, Attend and Tell~\cite{xu2015show}, which introduced attention mechanisms allowing the system to focus on specific image regions. Although these early practices established the foundation for cross-modal learning, they were limited by their dependence on supervised datasets and their inability to generalize across tasks without retraining.
A major step in vision-language learning were contrastive learning–based models like CLIP (Contrastive Language–Image Pretraining)~\cite{radford2021learning}, which jointly trained image and text encoders using large-scale image-text pairs. 
Building on earlier models like CLIP, VideoCLIP\cite{hu2021videoclip} applied the idea of contrastive learning to video. It used video and text pairs to train a model that could link motion in videos with language, using features that captured movement and appearance. 
ViCLIP~\cite{wang2024internvid} extended this line of work by introducing a contrastive loss tailored for video data. This allowed the model to better track how visual content changes over time, improving its ability to align video clips with text in a temporally consistent way. 
LLaVA\cite{liu2023llava} was one of the first models to link a vision encoder with an instruction-following language model. It connected the visual encoder from CLIP to the language model Vicuna using a small projection layer. This allowed the model to follow multi-modal instructions with only a few changes to the architecture. 
Video-LLaVA~\cite{lin2023video} used image-video pre-alignment which allows the LLM to learn a single unified visual representation. 
These earlier VLMs were limited in their ability to process long videos and capture fine-grained actions due to constraints in temporal modeling and input length. To overcome these limitations, specialized architectures were developed to support longer video sequences and more detailed action understanding. 
InternVideo2.5~\cite{Wang2025InternVideo25EV} introduced hierarchical temporal modeling and could process sequences of over 10,000 frames. A task preference optimization strategy was used to balance performance across different video understanding tasks. PerceptionLM~\cite{cho2025perceptionlm} focused on transparency and detailed understanding by training on 2.8 million human-labeled question-answer pairs and spatio-temporal captions.

More recent developments pushed VLMs beyond RGB-only processing to handle multiple sensory inputs simultaneously, which is relevant for datasets like Drive\&Act~\cite{martin2019drive_and_act_2019_iccv} that provide RGB, depth, infrared, and skeletal data. 
ImageBind~\cite{girdhar2023imagebind} used contrastive learning to align six modalities, which are images, text, audio, depth, thermal, and IMU data, in a shared image-bind embedding space.
LanguageBind~\cite{zhu2024languagebind} trained separate encoders for each modality but aligned them all to a shared language embedding space.
However, when using language modality on driver activity recognition former approaches like CMVRA~\cite{Lerch_2025_ICCV} fall short due to a lack of data. Therefore we provide a driver description dataset to let VLMs learn driver activity descriptions.

\subsection{Driver Activity Recognition Datasets}

Driver Activity Recognition (DAR) datasets were developed to support in-cabin driver behavior detection and classification. These datasets differ from general human activity recognition datasets in their focus on seated activities within vehicle cabins and their emphasis on detecting distraction and unsafe behaviors.
Early DAR datasets consisted primarily of static images with coarse activity labels. The State Farm Distracted Driver Detection dataset~\cite{statefarmdistracteddriverdetection} contains approximately 22,000 dashboard camera images labeled into 10 broad classes, for example ‘texting – right’, ‘operating radio’, ‘drinking’ etc. Similarly, the AUC Distracted Driver V1 dataset~\cite{abouelnaga2018realtime} provides roughly 13,000 images across similar categories. Both datasets use single-frame snapshots rather than video sequences.
Several multi-modal video‑based driver activity recognition datasets have been developed to overcome limitations in purely temporal or unimodal modeling. One example is DMD~\cite{ortega2020dmd} (Driver Monitoring Dataset), which comprises 41 hours of synchronized RGB, depth, and infrared (IR) video from 3 camera views (face, body, hands), collected from 37 drivers in both real‑vehicle and driving simulator settings. The recordings include a rich set of driver states and behaviors. 
Another notable multi-modal dataset is 3MDAD~\cite{Jegham2020_3MDAD} (Multiview, Multimodal, Multispectral Driver Action Dataset), which was collected from over 60 drivers performing 16 in-vehicle actions under both daytime and night time conditions. 
Despite the availability of multi-modal data and longer video sequences, the activity annotations in many driver datasets remain coarse. 
The Drive\&Act dataset~\cite{martin2019drive_and_act_2019_iccv} was introduced to address this limitation with a focus on fine-grained driver activity recognition using multi-view and multi-modal RGB-D recordings. It captures driver behavior from synchronized cameras targeting the face, body, and hands, enabling detailed observation of in-vehicle interactions. Unlike earlier datasets, Drive\&Act provides annotations at three levels: high-level tasks, fine-grained activities, and atomic action units with precise temporal boundaries. This is why we use the Drive\&Act atomic action units to generate fine-grained driver activity descriptions. 
Due to its strong class imbalance, the Drive\&Act dataset is a challenging dataset~\cite{Lerch2025self}. The performance of SotA VLMs on this dataset in section~\ref{sec:results} underscore this assessment.
Cañas et al.~\cite{Canas2025ExplorationVLMsDMS} evaluated VLMs for DMS using prompting on DMD. Unlike this prior exploratory work, we introduce a fine-grained natural-language description dataset to train and evaluate VLMs for in-cabin activity understanding.
\section{Proposed Method}

\begin{figure}[t]
  \centering
  \includegraphics[width=0.9\linewidth]{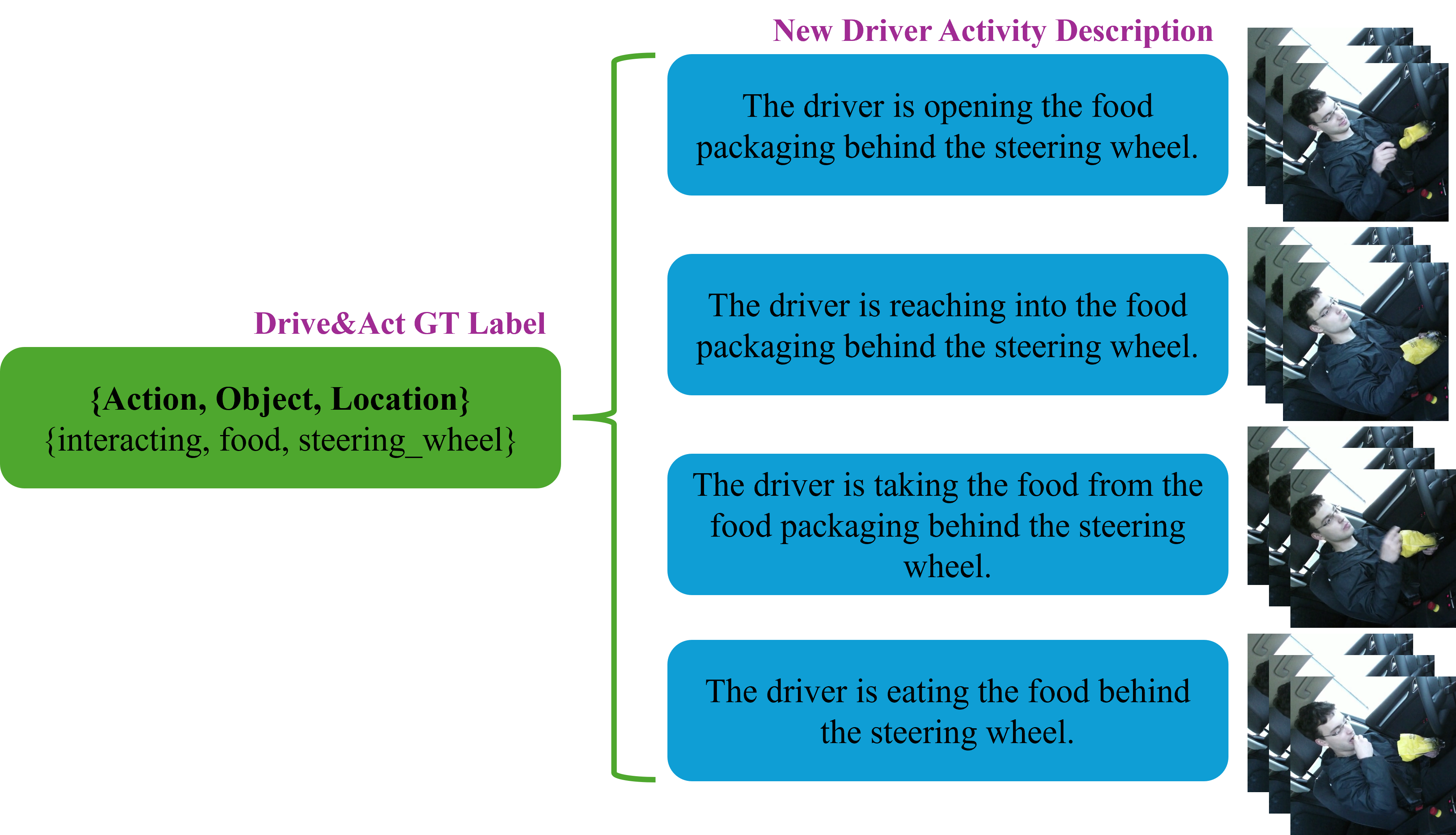}
   \caption{Principle of description generation based on Drive\&Act ground truth labels.}
   \label{fig:data}
\end{figure}

\begin{figure*}[t]
  \centering
  \includegraphics[width=0.9\linewidth]{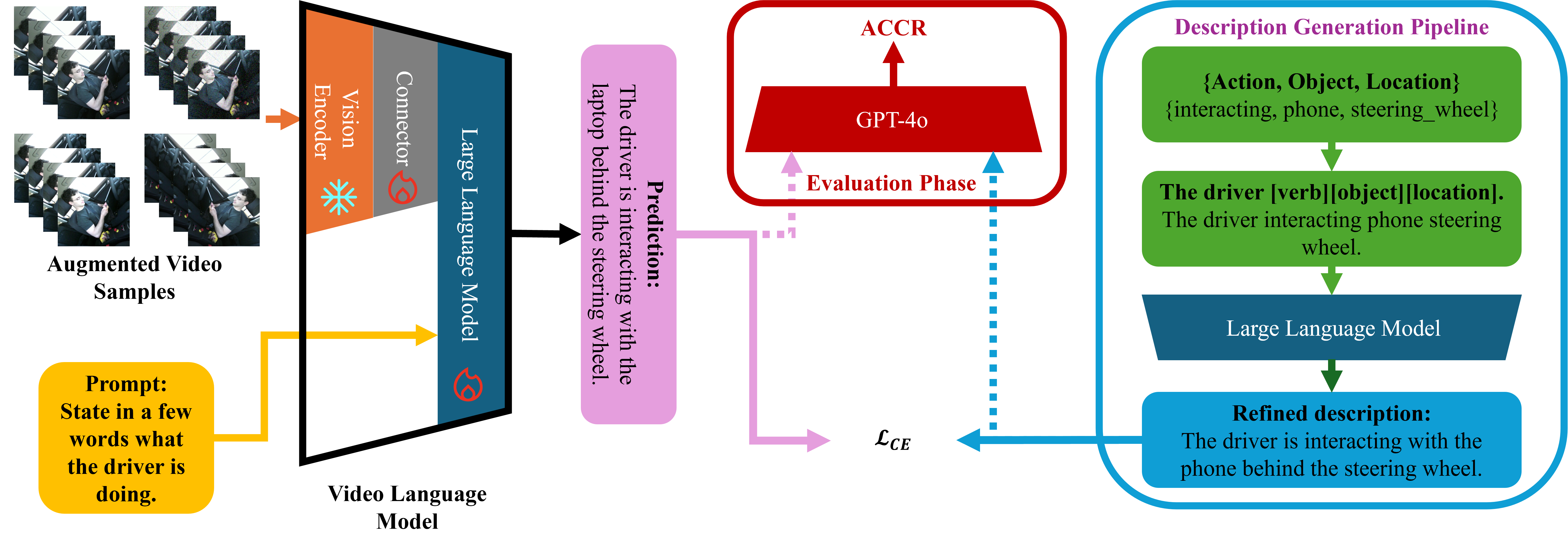}
   \caption{Proposed framework of training VLMs on driver actions. On the right side we show the flow of our automatic description generation for driver actions (green). During training we use the generated descriptions as ground truth. We train a video-language model to predict the description based on input videos and a constant prompt. During evaluation we use GPT-4o to evaluate the predicted descriptions using the ACCR metric.}
   \label{fig:principle}
\end{figure*}

\label{sec:method}
Our aim is to clear the path for training VLMs on driver distraction datasets. For the scope of this work we focus on creating fine-grained driver action descriptions for training VLMs (see Figure~\ref{fig:principle}) using the low-level atomic action units provided in the Drive\&Act dataset. It should be noted that the term ‘fine-grained description’ of driver action described in this work is different from Drive\&Act mid-level fine-grained labels, and will be used consistently throughout this work.

\subsection{Dataset Annotation}
\label{sec:dataannotation}
The Drive\&Act dataset defines each atomic action as a triplet consisting of:
\{Action, Object, Location\}.
These triplets describe the basic components of driver-object interactions, but they do not read like natural language. To make them more useful for training VLMs, each triplet was converted into a natural language sentence. A simple template was used for this conversion:
“The driver [verb] [object] [location].”
For example:
\begin{enumerate}
    \item \{retracting\_from, bottle, codriver\_footwell\} $\rightarrow$ The driver is retracting the bottle from the co-driver’s footwell.
    \item \{placing\_moving\_to, food, center\_console\_back\} $\rightarrow$ The driver is placing food in the center console.
\end{enumerate}

After automatically drafting these sentences, ChatGPT-4o was used to refine them for fluency and grammatical accuracy. The result was a set of fine-grained natural language descriptions that maintained atomic-level details but could also be used directly for model training. The hallucinations and inaccuracies in the generated descriptions were manually corrected.
Next, groups of these fine-grained descriptions were mapped to higher-level classes in the Drive\&Act dataset. E.g. the higher-level class 'interacting\_with\_phone' consists of the following atomic action triplets:

\begin{enumerate}
    \item \{reaching\_for, phone, trouser\_pocket\}
    \item \{retracting\_from, phone, trouser\_pocket\}
    \item \{interacting, phone, steering\_wheel\}
    \item \{placing\_moving\_to, phone, center\_console\_back\}
    \item \{retracting\_from, no\_object, center\_console\_back\}
\end{enumerate}

In some cases, the triplets were too vague to describe the driver’s action clearly. For example, \{interacting, food, front\_area\} appeared often but did not specify what kind of interaction took place. Depending on the video, this could mean opening the packaging of the food, reaching inside it, taking food out, or eating. To resolve such ambiguity, the related video clips were manually reviewed and annotated. The same triplet \{interacting, food, steering\_wheel\} was then rewritten into several fine-grained descriptions that matched the observed actions (see Figure~\ref{fig:data}).

In other cases, further fine-grained distinctions needed to be made. Although both newspaper and magazine are reading materials and share very similar fine-grained descriptions, it was important to maintain a clear distinction between the two classes. During the annotation process, the atomic triplets \{interacting, newspaper, ...\} and \{interacting, magazine, ...\} initially produced overlapping descriptions such as “The driver is reading the newspaper on the co-driver’s seat” or “The driver is reading the magazine on their lap.” Because of the visual similarity in these two actions, early training experiments showed that the model tended to overfit. Moreover, the model also failed to distinguish these two classes. To address this, video segments were manually reviewed and annotated with action-specific descriptions that emphasized contextual cues distinguishing the two objects.

Each long video in the Drive\&Act dataset was divided into shorter clips, each representing a logical sequence of fine-grained actions. For example, a sequence where the driver picks up a magazine, flips through it, reads, and places it down was labeled with the high-level description
“interacting with magazine.” Because only one object (the magazine) was involved, this clip was categorized as a Single Action.
In contrast, when two or more activities occurred in the same clip, for example, actions related
to handling a laptop like picking it up, retracting, and interacting with it, were followed by a brief
interruption in which the phone was reached for and used, after which the phone was placed down, and the laptop was finally placed in the co-driver’s footwell. Because more than one action type was present, the resulting clip was categorized as an overlapped action.

The Drive\&Act annotations contained 372 unique atomic action units. Our final dataset contains 460 unique fine-grained atomic action descriptions, an increase of 88 descriptions (+23.7\%). Our dataset splits ambiguous terms (e.g. “interacting with laptop”) into more precise descriptions (e.g. “opening laptop”, “working on laptop”, “closing laptop”).

\subsection{Data Pipeline for training}
This section describes the data preparation pipeline developed training on the modified Drive\&Act dataset. The pipeline was designed to handle data imbalance, generate sufficient training samples, and produce variations of clips that improve model generalization.
The Drive\&Act dataset contains uneven numbers of clips per activity class. For example, classes such as interacting with multimedia display have many clips, while others, like fastening seatbelt, are underrepresented. To train the model effectively, it is necessary to balance the dataset so that all classes have a similar number of clips across the splits (train, validation, and test).

In the imbalanced data, “interacting with multimedia display” had 295 clips, while “interacting with magazine” had only 13 clips. A mean value was calculated for the split, which came to around ~60 clips per class. This value was then used as the balancing target.
Classes above the value were reduced. For instance, the 295 clips of interacting with multimedia display were down sampled to 60 clips, and the 162 clips of pressing automation button were also reduced to 60 clips.
Classes below the value were increased using recycling and augmentation. For example, interacting with magazine had only 13 clips, and interacting with newspaper had 18 clips. New samples were created using recycling or augmentation until both classes reached 60 clips.
For classes that fell below the target value, additional samples had to be created. This was done in two stages. First, new clips were produced by recycling segments from overlapping videos belonging to the same class. If recycling alone did not provide enough material to reach the required count, further augmentation techniques were applied to expand the class.

\paragraph*{Recycling} was applied when overlapped-action clips contained relevant sub-sequences that could be extracted and reused for a specific activity class. For example, an overlapped sequence where a driver uses a laptop and then briefly checks the phone can provide training clips for both interacting with laptop and interacting with phone.
When clips were recycled, the absolute timestamps from the original video were removed and each retained segment was shifted to start at 0 s in the new clip. The durations of each action stayed identical, so only the time origin was changed.

\paragraph*{Video Augmentation} is used if recycling does not produce enough additional samples to bring a class up to the balancing value. In this stage, the remaining deficit was filled by creating new variants of existing clips. The augmentation included changes such as altering brightness or contrast, applying blur or noise, and flipping the frames horizontally. These transformations preserved the underlying action while introducing visual diversity, ensuring that rare classes could be expanded without losing their original meaning.
The pixel values of frames were scaled and shifted to vary brightness and contrast. Contrast factor sampled from $0.7$ to $1.6$, and Brightness offset sampled from $–40$ to $+40$.
Random Gaussian noise was added pixel-wise. Standard deviation sampled between $12.0$ and $25.0$.
Gaussian blurring was applied across the full frame. Kernel sizes randomly selected from $9$, $13$, $17$, or $21$. Larger kernels produced stronger blur, simulating camera defocus or motion blur.
Each frame was randomly mirrored along the vertical axis.
Frames were rotated with small angles around the center. Rotation angle sampled between $–15^\circ$ and $+15^\circ$.
Frames were resized to simulate zooming in or out. Scaling factor sampled between $0.85$ and $1.15$. When scaling up ($>1.0$), the image was cropped to the original size; when scaling down ($<1.0$), padding was added.
In some classes, even after recycling and applying augmentations, the number of samples still fell short of the target. In these cases, multiple augmentations could be applied in combinations. For example, one clip might combine blurring and rotation in order to generate more video clips until the target number for balancing the data is reached.

\section{Experiments}
\label{sec:results}
\subsection{Datasets}
The Drive\&Act dataset~\cite{martin2019drive_and_act_2019_iccv} encompasses 34 fine-grained and 12 high-level activities. We create the Drive\&Act description dataset from the more than 200 different action triplets~(see section~\ref{sec:dataannotation}). We use the Drive\&Act description dataset for all fine-tuning of VLMs.
DMD~(Driver Monitoring Dataset)~\cite{ortega2020dmd} comprises 41 hours of synchronized RGB, depth, and infrared video from 3 camera views~(face, body, hands), collected from 37 drivers in both real‑vehicle and driving simulator settings. The recordings include a rich set of driver states and behaviors. A derived subset, dBehaviourMD~(dBMD), includes annotations for 13 distraction actions.
Despite the availability of multi-modal data and longer video sequences, the activity annotations in many driver datasets remain coarse. DMD has 13 broad distraction category labels without marking finer temporal boundaries or decomposing each action into sub‑actions. This is valid for many driver distraction datasets such as StateFarm~\cite{statefarm2016} or 3DMD~\cite{Jegham2020_3MDAD}. In these datasets, an annotation like “drinking” does not tell how the driver reached for the bottle, opened it, took a sip, closed it, or placed it back, each of which could be separate atomic actions with their own time intervals.

\subsection{Models}
Given that Drive\&Act is structured around fine-grained human action recognition, models pretrained on large-scale video-text corpora or human activity datasets were prioritized. Such pretraining enables better generalization to subtle, temporally grounded driver actions.
The selection of models was guided by the goal of generating fine-grained, context-aware descriptions of in-cabin driver actions. At the outset, a survey of state-of-the-art VLMs was conducted, and three models were selected for detailed evaluation: Video-LLaVA~\cite{lin2023video}, InternVideo2.5~\cite{Wang2025InternVideo25EV}, and Perception-LM~\cite{cho2025perceptionlm}.

Video-LLaVA~\cite{lin2023video} is a vision-language model designed to understand both images and videos. It extends the LLaVA~\cite{liu2023llava}
architecture by adding support for video inputs. The model focuses on video-based instruction-following and detailed video understanding using natural language. 
Video-LLaVA uses Vicuna-7B, a large language model based on LLaMA, for generating text outputs. This model handles all instruction-following and text generation tasks.
Video-LLaVA uses the LanguageBind~\cite{zhu2024languagebind} Vision Transformer as the visual encoder. LanguageBind is capable of handling different modalities, including video, image, audio, depth, and infrared frames.
A trainable projection matrix connects the visual encoder output to the language model’s embedding space. This allows the model to align visual and language features.


Perception-LM~\cite{cho2025perceptionlm} is a unified vision-language model designed to handle a wide range of perception tasks through language-based instruction. It is capable of processing images, videos, point clouds, and multi-modal combinations using a single model architecture. The main idea is to treat different perception tasks as language modeling problems.
Perception-LM is built on the language model backbone of LLaMA (7B or 13B variants) and is extended with a frozen vision encoder and a perception instruction tuning mechanism. 
The vision encoder is based on SigLIP~\cite{Zhai_2023_siglip}, used to extract visual features from images or video frames.
A Q-Former maps visual features into a form that can be understood by the language model.


InternVideo2.5~\cite{Wang2025InternVideo25EV} consists of three main modules: 
The vision encoder is based on InternViT\cite{wang2024internvid} architecture, which takes individual RGB video frames as input. 
The connector module is a two-layer MLP, which projects the vision encoder’s 1024-dimensional outputs into the 4096-dimensional embedding space used by the language model. 
The language model is InternLM2.5-7B, a decoder-only transformer 
with a token embedding size of 4096 and 
a maximum sequence length configured up to 32,000 tokens depending on system memory. 

\begin{table*}[t]
\centering
\caption{Benchmarking of Video-language models on the Drive\&Act dataset grouped by description granularity.}
\label{tab:benchmark}

\begin{tabular}{@{}l l ccc c ccccc@{}}
\toprule
\multirow{2}{*}{Dataset} &
\multirow{2}{*}{VLM} &
\multicolumn{3}{c}{BERT Score~\cite{zhang2020bertscore}} &
\multirow{2}{*}{CLAIR~\cite{chan2023clair}} &
\multicolumn{5}{c}{ACCR Score~\cite{tong2025gveval}} \\
\cmidrule(lr){3-5} \cmidrule(lr){7-11}
 &  & 
Precision & Recall & F1 &
 &
Acc. & Complete. & Concise. & Relevance & ACCR Mean \\
\midrule

\multirow{3}{*}{DAA~\cite{martin2019drive_and_act_2019_iccv} coarse}
 & \multirow{1}{*}{VideoLLaVA~\cite{lin2023video}}
 & 0.47 & 0.56 & 0.50 & 38.26 & 32.99 & 26.46 & 80.59 & 33.08 & 43.28 \\
\cmidrule(lr){2-11}

 & \multirow{1}{*}{PerceptionLM~\cite{cho2025perceptionlm}}
 & 0.64 & 0.62 & \textbf{0.63} & 50.04 & 45.58 & 43.76 & 95.20 & 51.11 & 58.91 \\
\cmidrule(lr){2-11}

 & \multirow{1}{*}{InternVideo2.5~\cite{Wang2025InternVideo25EV}}
 & 0.60 & 0.59 & 0.59 & \textbf{58.81} & 49.84 & 48.54 & 86.75 & 55.42 & \textbf{60.14} \\

\midrule

\multirow{3}{*}{DAA~\cite{martin2019drive_and_act_2019_iccv} fine}
 & \multirow{1}{*}{VideoLLaVA~\cite{lin2023video}}
 & 0.45 & 0.51 & 0.47 & 23.17 & 17.06 & 12.78 & 52.01 & 20.73 & 25.64 \\
\cmidrule(lr){2-11}

 & \multirow{1}{*}{PerceptionLM~\cite{cho2025perceptionlm}}
 & 0.49 & 0.54 & \textbf{0.51} & 43.87 & 43.28 & 32.29 & 52.36 & 53.49 & 45.36 \\
\cmidrule(lr){2-11}

 & \multirow{1}{*}{InternVideo2.5~\cite{Wang2025InternVideo25EV}}
& 0.51 & 0.51 & \textbf{0.51} & \textbf{48.21} & 45.78 & 32.16 & 62.15 & 55.45 & \textbf{48.88} \\

\bottomrule
\end{tabular}
\end{table*}

\subsection{Evaluation Metrics}
For our VLM benchmark we use three different metrics. We show that even for simple classifcation tasks advanced metrics are required to evaluate the VLM predictions.

The BERTScore~\cite{zhang2020bertscore} evaluates a generated sentence against a reference sentence by comparing contextual token embeddings rather than exact word overlap. Tokens from the candidate and the reference are embedded with a pretrained Transformer, pairwise cosine similarities are computed, and a greedy token‑to‑token matching is used to derive precision, recall, and F1 scores. This approach aims to measure semantic equivalence, including paraphrases.

CLAIR~\cite{chan2023clair} is a method that uses large language models (LLMs) to evaluate image captions without relying on engineered similarity metrics.
The evaluation is performed by inserting both the candidate and reference captions into a fixed text prompt. This prompt asks the model to rate the similarity on a scale from 0 to 100 and provide a brief explanation. The prompt is structured as a text completion task, and the model is expected to return a formatted output with two fields: a numeric ‘score’ and a textual ‘reason’.

G-VEval~\cite{tong2025gveval} is a large language model (LLM)-based framework designed to evaluate text outputs for video and captioning tasks in the dimensions Accuracy, Completeness, Conciseness and Relevance (ACCR). Therefore we refer to it as the ACCR score
The evaluation uses human-written reference captions and the model-generated captions. The LLM is prompted to reason about the relationship between the candidate and the reference step by step and then produce a numeric score between 0 and 100. 

\subsection{Experimental Setup}
\label{sec:exp_setup}
We use AdamW optimizer, lr warm-up to $1e^{-4}$ in the first epoch and cosine lr decay for 9 epochs.
We use mixed precision and LoRA fine-tuning.
The InternVideo2.5 model was fine-tuned on the RGB modality of the modified Drive\&Act dataset. Frame sampling was set to approximately 1 frame per second for all experiments. While this sampling rate is coarse for fine-grained action recognition, it enabled faster experimentation to identify optimal parameter settings.
In order to find the best freeze patterns for the models, we conducted experiments on a subset of Drive\&Act consisting of 1428 samples with 17 classes. We report the results in the ablation study~(see section~\ref{sec:ablations})

. 

\begin{table}[t]
\centering
\caption{Performance comparison of InternVideo2.5~\cite{Wang2025InternVideo25EV} on Drive\&Act and dBMD datasets under zero-shot setting and fine-tuned on our Drive\&Act descriptions. }
\label{tab:grouped_runs}

\begin{tabular}{@{}llccccc@{}}
\toprule
Dataset &
Train &
Acc. &
Compl. &
Conc. &
Rel. &
Mean \\
\midrule

\multirow{2}{*}{DAA~\cite{martin2019drive_and_act_2019_iccv} coarse}
 & zs & 49.84 & 48.54 & 86.75 & 55.42 & 60.14 \\
 & ft & 90.49 & 85.52 & 98.95 & 90.42 & \textbf{91.34} \\
\midrule

\multirow{2}{*}{DAA~\cite{martin2019drive_and_act_2019_iccv} fine}
 & zs & 45.78 & 32.16 & 62.15 & 55.45 & 48.88 \\
 & ft & 73.55 & 68.21 & 88.46 & 88.88 & \textbf{79.77} \\
\midrule

\multirow{2}{*}{dBMD~\cite{ortega2020dmd}}
 & zs  & 71.35 & 45.10 & 86.88 & 61.15 & 66.12 \\
 & ft* & 71.04 & 66.25 & 94.38 & 73.65 & \textbf{76.33} \\
\midrule
 
\multicolumn{7}{c}{*the model is fine-tuned on Drive\&Act descriptions} \\

\bottomrule
\end{tabular}
\end{table}

\section{Results}
\subsection{Video Language Model Benchmarking}
Table~\ref{tab:benchmark} summarizes the benchmarking outcomes of the evaluated VLMs on the Drive\&Act dataset across coarse and fine-grained action descriptions. The results reveal clear performance trends concerning semantic alignment, temporal grounding, and descriptive quality.

\paragraph{BERT Score.}
PerceptionLM consistently achieves the highest performance according to the BERT-based similarity metrics. For coarse-grained descriptions, PerceptionLM attains the best F1 score of $0.63$ at temperature $T = 0.0$, surpassing both InternVideo2.5 ($F1 = 0.59$) and VideoLLaVA ($F1 = 0.50$). 
However, for fine-grained descriptions, all models converge to nearly identical BERT F1 scores in the range of $0.47–0.51$, indicating no significant differences. 

\paragraph{CLAIR and ACCR Metrics.}
While PerceptionLM excels in lexical similarity, InternVideo2.5 demonstrates superior performance on interpretability-oriented measures, namely CLAIR and ACCR, which better evaluate narrative completeness and relevance. InternVideo2.5 achieves the highest CLAIR score ($58.81$) and the best mean ACCR score ($60.14$), compared to PerceptionLM ($58.91$) and VideoLLaVA ($43.28$). A similar pattern emerges in fine-grained descriptions, where InternVideo2.5 maintains an ACCR Mean of $48.88$, marginally outperforming PerceptionLM ($45.36$). These results suggest that InternVideo2.5 produces more coherent, complete, and contextually relevant video-text descriptions, particularly under moderate temperature conditions.

\paragraph{Overall Performance.}
Across all metrics, Video-LLaVA consistently yields the weakest results, with both CLAIR and ACCR scores significantly lower than the competing models, indicating limited temporal reasoning and semantic grounding. In contrast, PerceptionLM and InternVideo2.5 perform comparably on coarse tasks, though InternVideo2.5 demonstrates more stable and contextually aligned performance as description granularity increases.
Despite notable differences among the models, overall performance levels remain considerably lower than those observed in supervised or self-supervised classification settings. Even InternVideo2.5, the best-performing model, achieves mean ACCR scores below $61$, underscoring the persistent challenge of generating high-quality, semantically accurate video-language descriptions without task-specific finetuning. These findings emphasize the necessity of a dedicated training dataset to enhance fine-grained reasoning and improve generalization capabilities in open-ended video-language understanding.

\subsection{Fine-tuning and Generalization}

Table~\ref{tab:grouped_runs} presents the comparative performance of InternVideo2.5 under zero-shot (zs) and fine-tuned (ft) evaluation settings on the Drive\&Act and dBMD datasets. The results clearly demonstrate that fine-tuning on driver activity descriptions substantially enhances both descriptive quality and semantic alignment across all evaluation metrics.
For both coarse- and fine-grained descriptions, fine-tuning on the Drive\&Act description dataset leads to a pronounced improvement over zero-shot performance. On the coarse-grained level, mean ACCR increases from $60.14$ (zs) to $91.34$ (ft). Similarly, for fine-grained actions, mean ACCR improves from $48.88$ to $79.77$, marking a relative gain of more than 30 percentage points. These improvements indicate that model adaptation to the descriptive distribution of driver activities significantly enhances contextual understanding and coherence in generated outputs.
Although the fine-tuning procedure uses only Drive\&Act descriptions, performance on dBMD also improves notably. The mean ACCR rises from $66.12$ in the zero-shot setting to $76.33$ after fine-tuning, indicating positive transfer effects. This demonstrates that fine-tuning on one dataset of human actions in the driving domain can generalize effectively to distinct monitoring scenarios.

The observed improvements confirm the strong impact of domain-adaptive fine-tuning for multi-modal VLMs in structured domains such as in-cabin driver monitoring. While InternVideo2.5 exhibits reasonable descriptive capability in zero-shot mode, the magnitude of the performance gap underscores the necessity of supervised domain adaptation to achieve reliable activity interpretation. The positive transfer to dBMD further validates the proposed Drive\&Act description dataset as a beneficial resource for enhancing cross-domain generalization and contextual grounding in driver behavior understanding.

\subsection{Ablation Study}
\label{sec:ablations}
We evaluate different freezing patterns for the VLM components ViT, LLM and connector. For these experiments we use a subset of the original Drive\&Act dataset as described in section \ref{sec:exp_setup}. As shown in Table~\ref{tab:abl_finetuning}, the performance dropped drastically when ViT was trainable. Therefore we decided to freeze the ViT and make the LLM and connector trainable.

\subsection{Discussion}
\begin{table}[t]
\centering
\caption{Ablations on different freeze patterns for VLM fine-tuning. The models are fine-tuned on a subset~(see section~\ref{sec:exp_setup}) of DAA.}
\label{tab:abl_finetuning}

\begin{tabular}{@{}lcccc@{}}
\toprule
\multirow{2}{*}{Dataset} &
\multicolumn{3}{c}{Trainable} &
\multirow{2}{*}{ACCR Mean} \\
\cmidrule(lr){2-4}
 & ViT & Connect. & LLM & \\
 \midrule
\multirow{3}{*}{DAA~\cite{martin2019drive_and_act_2019_iccv} coarse}
 & \checkmark & \xmark & \checkmark & 36.03 \\
 & \xmark & \checkmark & \checkmark & \textbf{87.49} \\
 & \checkmark & \checkmark & \checkmark & 33.06 \\
\midrule
\multirow{3}{*}{DAA~\cite{martin2019drive_and_act_2019_iccv} fine}
 & \checkmark & \xmark & \checkmark & 28.54 \\
 & \xmark & \checkmark & \checkmark & \textbf{71.58} \\
 & \checkmark & \checkmark & \checkmark & 29.56\\
\bottomrule
\end{tabular}
\end{table}

The comprehensive evaluation across both benchmarking and fine-tuning experiments provides key insights into the current limitations and opportunities for video-language understanding in the driver activity domain. A limitation of this work is that CLAIR and ACCR rely on an LLM judge, which may introduce evaluator bias. These metrics were used because open-ended driver-action descriptions cannot be fully evaluated using lexical overlap alone. To improve transparency, ACCR was prompted to provide reasoning for each score, and a subset of score-reason pairs was manually inspected for plausibility. 
The scalability of this pipeline to larger datasets could be achieved by replacing manual correction with an ensemble of video-grounded VLMs and specialized pre-processing tools. The employment of pre-processing tools, such as object tracking and pose estimation, has the potential to furnish structured spatial and temporal context, thereby diminishing the necessity for manual review.

\paragraph{Zero-Shot Performance and Benchmark Limitations.}
The benchmarking study on the Drive\&Act dataset highlights notable weaknesses in existing large-scale VLMs when applied in a zero-shot setting. Across all evaluated models, absolute performance on all metrics remains relatively low. 
Such results indicate that these models, although trained on large and diverse video corpora, struggle with recognizing subtle distinctions between semantically similar fine-grained actions in in-cabin monitoring contexts. VideoLLaVA exhibits consistently poor scores across all metrics, revealing limited temporal reasoning and contextual grounding capabilities. These findings collectively underscore the need for task-oriented adaptation to achieve reliable semantic and temporal fidelity in human activity description tasks.

\paragraph{Impact of Fine-Tuning on the Drive\&Act Description Dataset.}
Fine-tuning InternVideo2.5 on our new Drive\&Act description dataset yields substantial performance gains across all ACCR metrics, clearly demonstrating the value of domain-specific description data. 
The performance improvements indicate that grounded training not only enhances model accuracy and completeness but also improves overall narrative coherence and relevance. 
The results underscore the relevance of fine-tuning of VLMS for applications in driver activity recognition systems.

\paragraph{Cross-Dataset Generalization.}
An encouraging outcome of this study is the observed generalization to an unseen dataset. 
The improvement on dBMD suggests that the linguistic structure and conceptual semantics learned from Drive~\&Act transfer effectively to related domains of driver monitoring, enabling contextual understanding beyond the training distribution. The results imply that descriptive fine-tuning fosters a robust video-language grounding generalizing across driver activity recognition datasets.

\paragraph{Implications and Future Directions.}
Taken together, the findings highlight two complementary insights: (1) large-scale video pretraining alone is insufficient to model the structured, goal-directed nature of in-cabin human behavior, and (2) introducing a targeted descriptive dataset meaningfully bridges this gap. The Drive\&Act description dataset thereby establishes the first resource enabling the training and evaluation of VLMs for driver activity reasoning through natural language representations. Future work should extend this dataset with additional annotator diversity, multi-turn temporal descriptions, and explicit state–intent annotations to further enhance semantic grounding and causal reasoning in human-vehicle interaction contexts.

\section{Conclusion}
\label{sec:conclusion}
This work presents the first natural language description dataset for driver activity understanding and provides a comprehensive analysis of current VLMs in this context. Our results reveal that existing VLMs exhibit limited zero-shot capability for interpreting fine-grained driver actions, emphasizing the domain gap between general video pretraining and specialized in-cabin activity understanding. By fine-tuning on the proposed Drive\&Act description dataset, we improve the ACCR score from $49$ to $80$ in descriptive quality of fine-grained driver activities. Further, we demonstrate that fine-tuned on our Drive\&Act description dataset we improve the ACCR score on the dBMD dataset from $66$ to $76$. These outcomes confirm the effectiveness of domain-adaptive description-based training for enhancing driver monitoring. Future research should build upon this foundation by expanding descriptive coverage, integrating temporal reasoning, and exploring multi-modal larger-scale, cross-domain datasets to further advance explainable and reliable driver activity understanding systems. Potential future topics may also include cognitive distraction~\cite{HamiltonGrabowski2013}.

\section*{Acknoledgement}
The work has been funded under the funding code 19A24002K by the Federal Ministry for Economic Affairs and Climate Action of Germany (BMWK) on the basis of a decision by the German Bundestag and by the European Union. The work was performed in the project SALSA (https://projekt-salsa.de/).

\bibliographystyle{IEEEtran}
\bibliography{egbib}

\end{document}